\documentclass{article}

\usepackage[preprint]{neurips_2021}

\usepackage[utf8]{inputenc} % allow utf-8 input
\usepackage[T1]{fontenc}    % use 8-bit T1 fonts
\usepackage{hyperref}       % hyperlinks
\usepackage{url}            % simple URL typesetting
\usepackage{booktabs}       % professional-quality tables
\usepackage{amsfonts}       % blackboard math symbols
\usepackage{nicefrac}       % compact symbols for 1/2, etc.
\usepackage{microtype}      % microtypography
\usepackage{xcolor}         % colors
\usepackage{graphicx}   
\usepackage{adjustbox}

\title{Classification of Computer Aided Engineering (CAE) Parts Using Graph Convolutional Networks}

\author{%
  Alok Warey \\
  Global Research and Development  \\
  General Motors\\
  Warren, MI 48090 \\

  \And
  Rajan Chakravarty \\
  Vehicle Optimization\\ 
  General Motors   \\
  Warren, MI 48090 \\
}

\begin{document}

\maketitle

\begin{abstract}
CAE engineers work with hundreds of parts spread across multiple body models. A Graph Convolutional Network (GCN) was used to develop a CAE parts classifier. As many as 866 distinct parts from a representative body model were used as training data. The parts were represented as a three-dimensional (3-D) Finite Element Analysis (FEA) mesh with values of each node in the x, y, z coordinate system. The GCN based classifier was compared to fully connected neural network and PointNet based models. Performance of the trained models was evaluated with a test set that included parts from the training data, but with additional holes, rotation, translation, mesh refinement/coarsening, variation of mesh schema, mirroring along x and y axes, variation of topographical features, and change in mesh node ordering. The trained GCN model was able to achieve 88.5\% classification accuracy on the test set i.e., it was able to find the correct matching part from the dataset of 866 parts despite significant variation from the baseline part. A CAE parts classifier demonstrated in this study could be very useful for engineers to filter through CAE parts spread across several body models to find parts that meet their requirements.
\end{abstract}

\section{Introduction}

The objective of this study was to develop a CAE parts classifier that could be used to identify similar parts from hundreds of CAE parts spread across multiple body models. As many as 866 distinct parts from a representative body model were used as training data. The parts were represented as a 3-D FEA mesh with values of each node in the x, y, z coordinate system. For all parts in this study, the x, y, z coordinate values of each mesh node were scaled i.e., zero centered and normalized by their respective mean and standard deviation. 

\section{Graph Convolutional Network}

Graph Convolutional Networks (GCNs) can work directly on graphs and take advantage of their structural information [1]. Graph convolution is mathematically defined as follows [1, 2]:

\begin{equation}
h_i^{(l+1)} = \sigma(b^{(l)} + \sum_{j\in\mathcal{N}(i)}\frac{1}{c_{ij}}h_j^{(l)}W^{(l)})         
\end{equation}

where  \begin{math} h_j^{(l)} \end{math} and \begin{math} h_i^{(l+1)} \end{math} are the input and output node features, \begin{math} \mathcal{N}(i) \end{math} is the set of neighbors of node \begin{math} i, c_{ij} \end{math} is the product of the square root of node degrees i.e., \begin{math} c_{ij} = \sqrt{|\mathcal{N}(i)|}\sqrt{|\mathcal{N}(j)|}\end{math}, and $\sigma$ is an activation function.

For this study, the mesh for each part was converted into an undirected graph with self-loops for each node. An example is shown in Figure 1. As stated before, node features were the scaled x, y, z coordinate values of each node in the mesh. No edge level features were used. The graphs were then passed through a GCN for graph classification as shown in Figure 2. The final/output layer of the network was of size (866, 1) with a Softmax activation function. The GCN model was developed using Deep Graph Library [2] with PyTorch backend [3]. 

\begin{figure}[h!]
  \centering
  \includegraphics[scale=0.34]{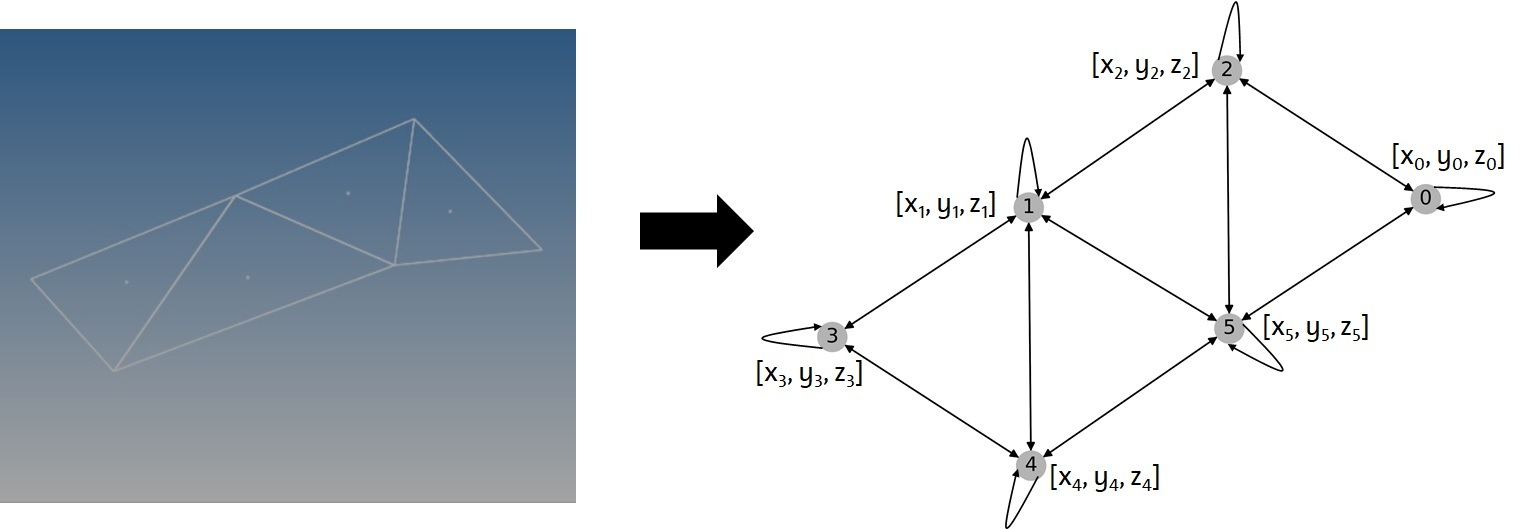}
  \caption{Example of a mesh converted into an undirected graph.}
\end{figure}

\begin{figure}[h!]
  \centering
  \includegraphics[scale=0.38]{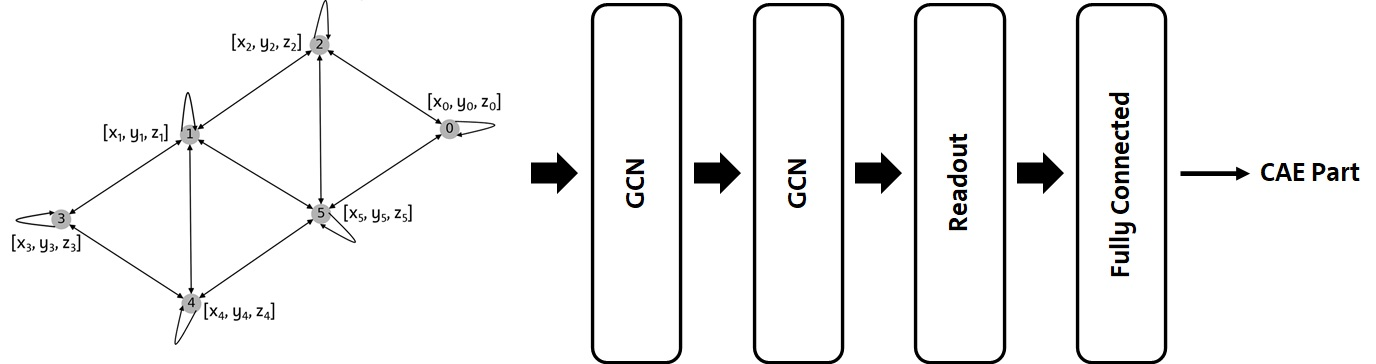}
  \caption{Graph Convolutional Network model.}
\end{figure}

\section{Fully Connected Neural Network}

A fully connected neural network model was developed to classify parts as shown in Figure 3. For this model, the three-dimensional mesh (n\textsubscript{nodes}, 3) was reshaped into an one-dimensional vector (n\textsubscript{nodes} x 3, 1). Since the parts were of widely varying sizes, zero padding was used to ensure that the size of the one-dimensional vector was the same for all parts. The size of the input layer of the model was equal to the length of the biggest part, when reshaped into a one-dimensional vector.

\begin{figure}[h!]
  \centering
  \includegraphics[scale=0.5]{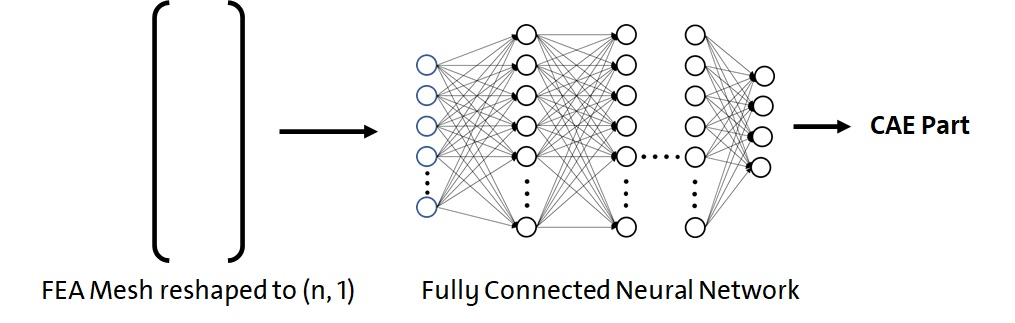}
  \caption{Fully Connected Neural Network model.}
\end{figure}

\section{PointNet}

PointNet is novel type of neural network that directly consumes point clouds [4]. PointNet, provides a unified architecture for applications ranging from object classification, part segmentation, to scene semantic parsing. The network learns a set of optimization functions/criteria that select interesting or informative points of the point cloud and encode the reason for their selection. The final fully connected layers of the network aggregate these learnt optimal values into a global descriptor for the entire shape (shape classification) or are used to predict per point labels (shape segmentation). The classification network takes n points as input, applies input and feature transformations, and then aggregates point features by max pooling. The output is classification score for m classes. Batchnormalization [5] is used for all layers with ReLU activation function [6]. Dropout layers [7] are used for the last couple layers in the classification net [4]. The PointNet architecture was adapted for the current study to accept the 3-D FEA mesh (n\textsubscript{nodes}, 3) as input as shown in Figure 4. Zero padding was again used to ensure that all parts had the same size i.e., n\textsubscript{nodes} was equal to the length of the biggest part in the dataset for all parts.

\begin{figure}[h!]
  \centering
  \includegraphics[scale=0.5]{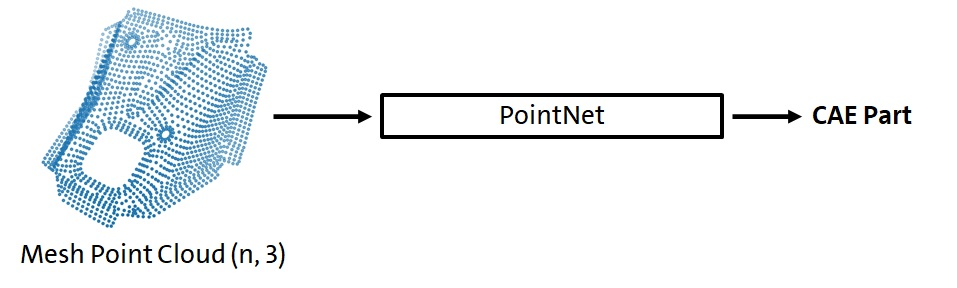}
  \caption{PointNet model.}
\end{figure}

\section{Results}

Performance of the trained models was evaluated with a test set that included parts with:

• Additional holes \\
• Rotation and translation \\
• Mesh refinement/coarsening \\
• Variation of mesh schema \\
• Mirroring along x and y axes \\
• Variation of topographical features 

A big advantage of GCN and PointNet models over fully connected neural networks is that they are permutation invariant i.e., they do not depend on the arbitrary ordering of the nodes in the mesh. This was demonstrated by changing the ordering of the mesh nodes for a sample part before using the three trained models. The test set was divided into 5 subsets (A to E). Model performance results for each subset are given below.

\subsection{Test Subset A}

This subset included translated, rotated and rotated+translated versions of a part labelled "P2" shown in Figure 5, which was randomly selected from the 866 parts dataset. Model performance results on this test subset are given in Table 1 i.e., whether the models were correctly able to classify the part. Both PointNet and GCN models misclassified with translation and rotation of the part. 

\begin{figure}[h!]
  \centering
  \includegraphics[scale=0.5]{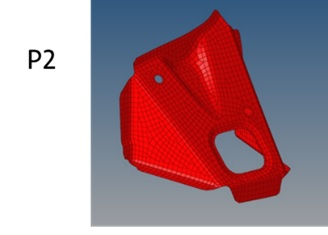}
  \caption{Part P2 randomly selected from the 866 parts training dataset.}
\end{figure}

\begin{table}[ht]
\caption{Model performance on Test Subset A} 
\centering
\begin{adjustbox}{width=1\textwidth}
\small
\begin{tabular}{cccc}
  \hline
 Test Part & Fully Connected Neural Network & PointNet & Graph Convolutional Network (GCN)  \\ 
  \hline
  P2 Translated & $\checkmark$ & $\times$ & $\checkmark$  \\ 
    P2 Rotated & $\checkmark$ & $\checkmark$ & $\times$  \\ 
  P2 Rotated + Translated & $\checkmark$ & $\times$ & $\times$   \\ 
   \hline
\end{tabular}
\end{adjustbox}
\end{table} 

\subsection{Test Subset B}

This subset included variations on part P2 such as addition of 1-5 holes, finer and coarser mesh, and change in the mesh schema from quadrilateral to triangular elements as shown in Figure 6. Model performance results on this test subset are given in Table 2.

\begin{figure}[h!]
  \centering
  \includegraphics[scale=0.34]{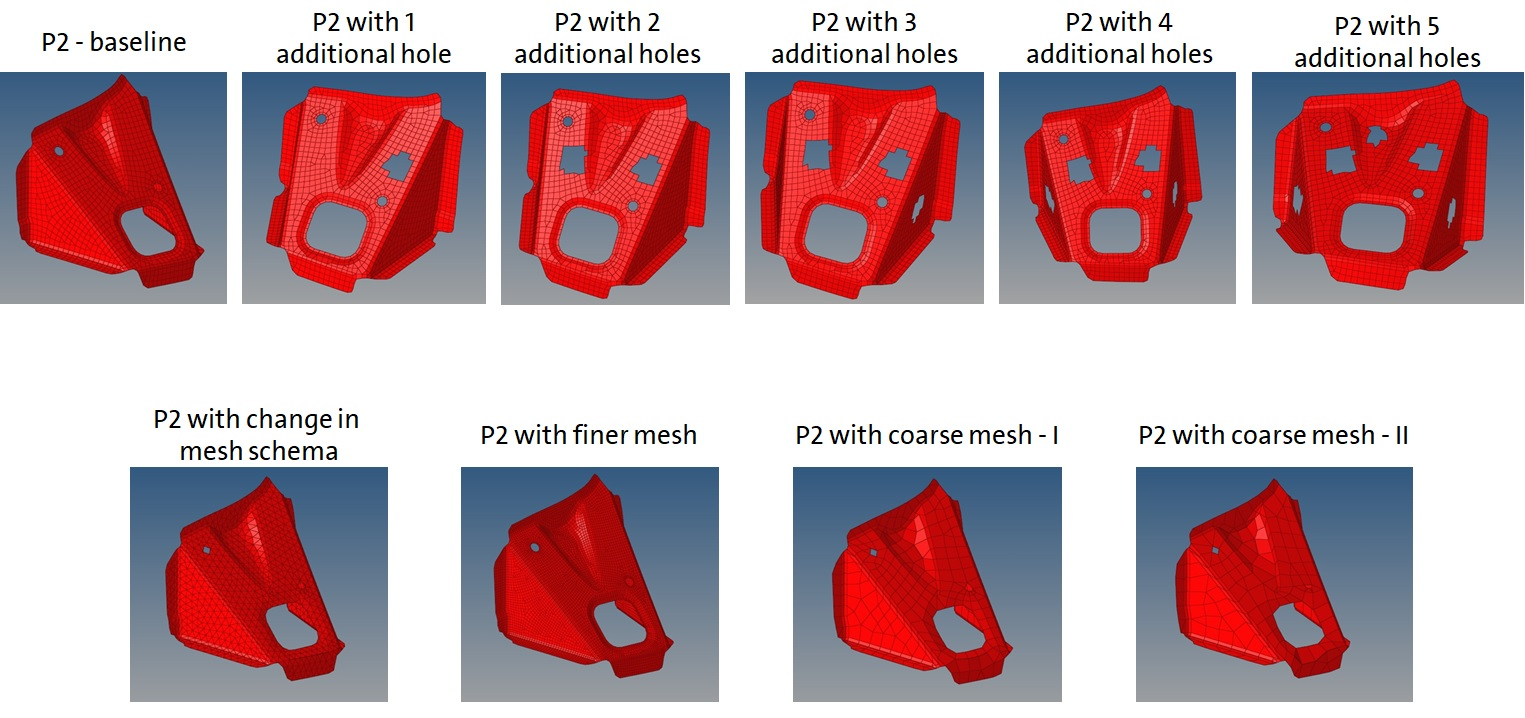}
  \caption{Images of parts in Test Subset B.}
\end{figure}

\begin{table}[h!]
\caption{Model performance on Test Subset B} 
\centering
\begin{adjustbox}{width=1\textwidth}
\small
\begin{tabular}{cccc}
  \hline
 Test Part & Fully Connected Neural Network & PointNet & Graph Convolutional Network (GCN)  \\ 
  \hline
  P2 with 1 additional hole & $\checkmark$ & $\checkmark$ & $\checkmark$  \\ 
  P2 with 2 additional holes & $\checkmark$ & $\checkmark$ & $\checkmark$  \\ 
  P2 with 3 additional holes & $\checkmark$ & $\checkmark$ & $\checkmark$  \\ 
  P2 with 4 additional holes & $\checkmark$ & $\checkmark$ & $\checkmark$  \\ 
  P2 with 5 additional holes & $\checkmark$ & $\checkmark$ & $\checkmark$  \\ 
  P2 with change in mesh schema & $\times$ & $\checkmark$ & $\checkmark$  \\ 
  P2 with finer mesh & $\checkmark$ & $\checkmark$ & $\checkmark$  \\ 
  P2 with coarse mesh - I & $\times$ & $\checkmark$ & $\checkmark$  \\ 
  P2 with coarse mesh - II & $\times$ & $\checkmark$ & $\checkmark$  \\ 
   \hline
\end{tabular}
\end{adjustbox}
\end{table} 

\subsection{Test Subset C}

This subset included variations on part P2 such as change in the mesh schema and mirrored versions along the x and y axes as shown in Figure 7. Model performance results on this test subset are given in Table 3.

\begin{figure}[h!]
  \centering
  \includegraphics[scale=0.4]{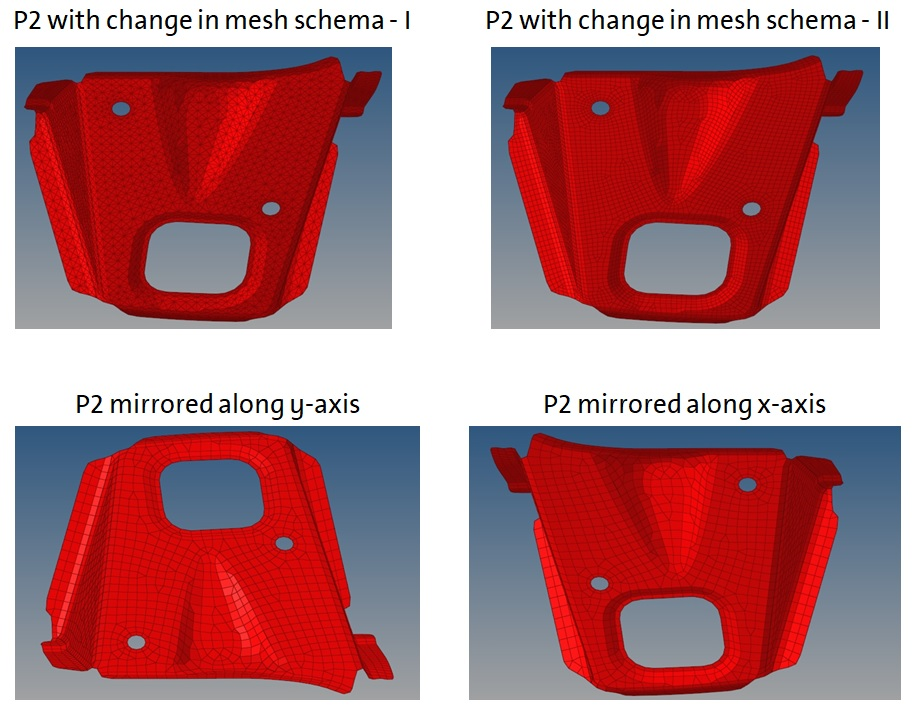}
  \caption{Images of parts in Test Subset C.}
\end{figure}

\begin{table}[h!]
\caption{Model performance on Test Subset C} 
\centering
\begin{adjustbox}{width=1\textwidth}
\small
\begin{tabular}{cccc}
  \hline
 Test Part & Fully Connected Neural Network & PointNet & Graph Convolutional Network (GCN)  \\ 
  \hline
  P2 with change in mesh schema - I & $\checkmark$ & $\checkmark$ & $\checkmark$  \\ 
  P2 with change in mesh schema - II & $\checkmark$ & $\checkmark$ & $\checkmark$  \\ 
  P2 mirrored along y-axis & $\checkmark$ & $\times$ & $\times$  \\ 
  P2 mirrored along x-axis & $\checkmark$ & $\times$ & $\checkmark$  \\ 
   \hline
\end{tabular}
\end{adjustbox}
\end{table} 

\subsection{Test Subset D}

This subset included 5-15\% scale up of the part P2 and variations in topographical features as shown in Figure 8. Model performance results on this test subset are given in Table 4.

\begin{figure}[h!]
  \centering
  \includegraphics[scale=0.3]{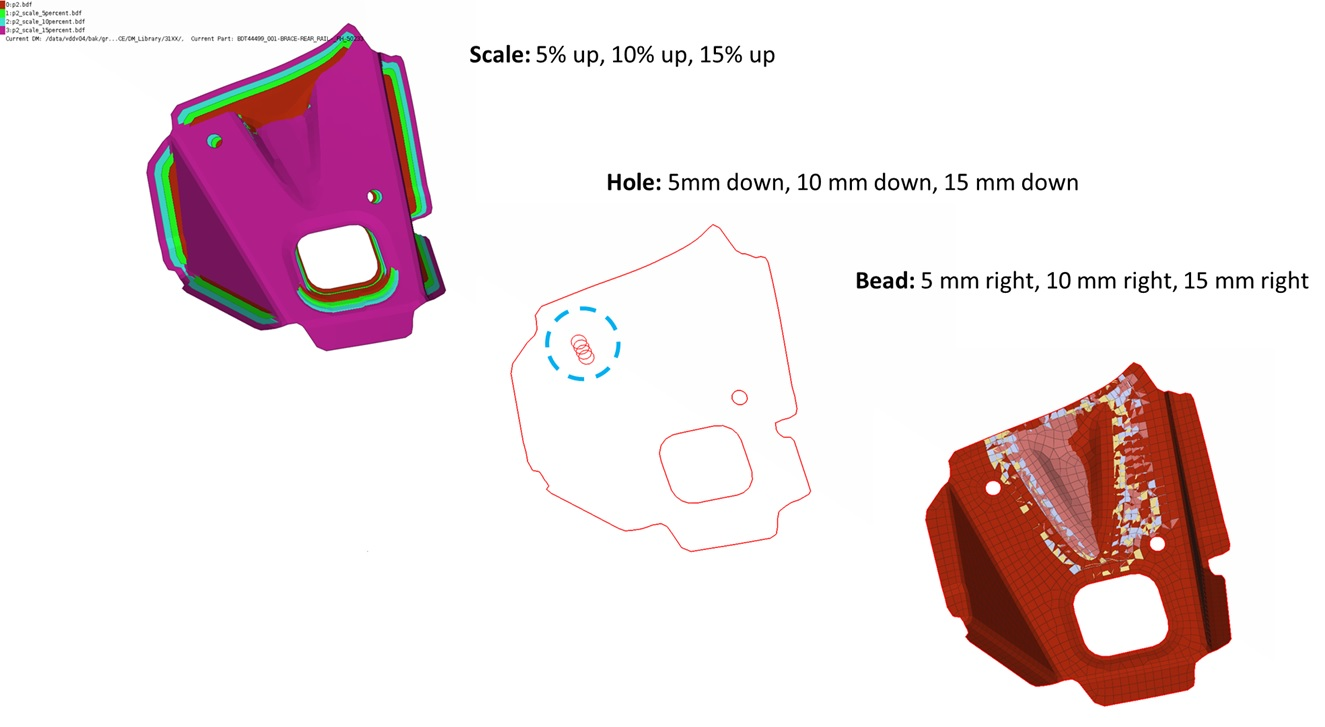}
  \caption{Images of parts in Test Subset D.}
\end{figure}

\begin{table}[h!]
\caption{Model performance on Test Subset D} 
\centering
\begin{adjustbox}{width=1\textwidth}
\small
\begin{tabular}{cccc}
  \hline
 Test Part & Fully Connected Neural Network & PointNet & Graph Convolutional Network (GCN)  \\ 
  \hline
  P2 scale up by 5\% & $\checkmark$ & $\checkmark$ & $\checkmark$  \\ 
  P2 scale up by 10\% & $\checkmark$ & $\checkmark$ & $\checkmark$  \\ 
  P2 scale up by 15\% & $\checkmark$ & $\checkmark$ & $\checkmark$  \\ 
  P2 with 5mm hole shift & $\checkmark$ & $\checkmark$ & $\checkmark$  \\ 
  P2 with 10mm hole shift & $\checkmark$ & $\checkmark$ & $\checkmark$  \\ 
  P2 with 15mm hole shift & $\checkmark$ & $\checkmark$ & $\checkmark$  \\ 
  P2 with 5mm bead shift & $\checkmark$ & $\checkmark$ & $\checkmark$  \\ 
  P2 with 10mm bead shift & $\checkmark$ & $\checkmark$ & $\checkmark$  \\ 
  P2 with 15mm bead shift & $\checkmark$ & $\checkmark$ & $\checkmark$  \\ 
   \hline
\end{tabular}
\end{adjustbox}
\end{table} 

\subsection{Test Subset E}

This subset included a part randomly selected from the 866 parts dataset. The ordering of the nodes in the mesh was arbitrarily changed to check for permutation invariance of the models. Unlike the fully connected neural network, both the GCN and PointNet models were still able to correctly classify the part as given in Table 5.

\begin{table}[h!]
\caption{Model performance on Test Subset E} 
\centering
\begin{adjustbox}{width=1\textwidth}
\small
\begin{tabular}{cccc}
  \hline
 Test Part & Fully Connected Neural Network & PointNet & Graph Convolutional Network (GCN)  \\ 
  \hline
  Change in mesh node ordering & $\times$ & $\checkmark$ & $\checkmark$  \\ 
   \hline
\end{tabular}
\end{adjustbox}
\end{table}

\section{Summary}

A Graph Convolutional Network (GCN) was used to develop a CAE parts classifier. As many as 866 distinct parts from a representative body model were used as training data. The parts were represented as a three-dimensional (3-D) Finite Element Analysis (FEA) mesh with values of each node in the x, y, z coordinate system. The GCN based classifier was compared to fully connected neural network and PointNet based models. Performance of the trained models was evaluated with a test set that included parts from the training data, but with additional holes, rotation, translation, mesh refinement/coarsening, variation of mesh schema, mirroring along x and y axes, variation of topographical features, and change in mesh node ordering. The trained GCN model was able to achieve 88.5\% classification accuracy on the test set i.e., it was able to find the correct matching part from the dataset of 866 parts despite significant variation from the baseline part. By comparison, the fully connected neural network and PointNet models achieved 84.6\% classification accuracy on the test set. Prediction accuracy of all models can be further improved by including examples of these variations from the baseline parts in the training data. 

\section*{References}
\medskip
\small

[1] T. N. Kipf, and M. Welling, Semi-Supervised Classification with Graph Convolutional Networks, 2016, {\it arXiv:1609.02907}.

[2] M. Wang, D. Zheng, Z. Ye, Q. Gan, M. Li, X. Song, J. Zhou, C. Ma, L. Yu, Y. Gai, T. Xiao, T. He, G. Karypis, J. Li, and Z. Zhang, Deep Graph Library: A Graph-Centric, Highly-Performant Package for Graph Neural Networks, 2019, {\it arXiv:1909.01315}.

[3] A. Paszke, S. Gross, F. Massa, A. Lerer, J. Bradbury, G. Chanan, T. Killeen, Z. Lin, N. Gimelshein, L. Antiga, A. Desmaison, A. Köpf, E. Yang, Z. DeVito, M. Raison, A. Tejani, S. Chilamkurthy, B. Steiner, L. Fang, J. Bai, and S. Chintala, PyTorch: An Imperative Style, High-Performance Deep Learning Library, 2019, {\it arXiv:1912.01703}.

[4] C. R. Qi, H. Su, K. Mo, and L. J. Guibas, PointNet: Deep Learning on Point Sets for 3D Classification and Segmentation, 2018, {\it arXiv:1612.00593v2}.

[5] S. Ioffe, and C. Szegedy, Batch Normalization: Accelerating Deep Network Training by Reducing Internal Covariate Shift, 2015, {\it arXiv:1502.0316}.

[6] V. Nair, and G. E. Hinton, Rectified Linear Units Improve Restricted Boltzmann Machines, Proceedings of the 27th International Conference on Machine Learning, 2010, {\it https://doi.org/10.5555/3104322.3104425}.  

[7] N. Srivastava, G. Hinton, A. Krizhevsky, I. Sutskever, and R. Salakhutdinov, Dropout: A Simple Way to Prevent Neural Networks from Overfitting, 2014, {\it Journal of Machine Learning Research} {\bf 15}:1929-1958.

%%%%%%%%%%%%%%%%%%%%%%%%%%%%%%%%%%%%%%%%%%%%%%%%%%%%%%%%%%%%

\end{document}